\documentclass[runningheads]{llncs}

\usepackage{eccv}

\usepackage{eccvabbrv}      
\usepackage{graphicx}
\usepackage{booktabs}
\usepackage{multirow}
\usepackage{amsmath,amssymb}
\usepackage[accsupp]{axessibility}

\usepackage{hyperref}

\usepackage{orcidlink}

\begin{document}

\title{Bridging Semantics and Physics with Constrained LLMs for Safe and Trustworthy Robotic Manipulation}

\titlerunning{Constrained LLMs for Safe and Trustworthy Manipulation}

\author{
Wenhao Hong\inst{1} \and
Lan Wei\inst{1} \and
Dandan Zhang\inst{1}\thanks{Corresponding author.}
}

\authorrunning{W. Hong et al.}

\institute{
Imperial College London, London, UK\\
\email{d.zhang17@imperial.ac.uk}
}

\maketitle

\begin{abstract}
A language-guided robot operating in a real kitchen must do more than produce a plan that appears correct. It must also execute that plan safely in cluttered environments under imperfect perception.
Large language models (LLM) can decompose instructions into action sequences, yet a language-action gap remains: a plan may appear valid linguistically while being physically infeasible under kinematic and collision constraints.
We bridge this gap by formalizing the reasoning-execution boundary as a typed contract.
From RGB-D observations, the system grounds perceived objects in an explicit, collision-aware scene model and constrains language-level decisions through schema-validated tool calls defined by the Model Context Protocol (MCP), rejecting malformed commands before they reach the robot.
Each validated call is deterministically grounded in a MoveIt Task Constructor pipeline, where candidate motions are evaluated against the reconstructed planning scene in a verify-then-act step.
Only trajectories that pass both kinematic and collision checks are sent to the robot.
On a physical UFactory 850, the method achieves up to 80\% success across ten trials per task on pouring tasks involving liquids, granular media, and discrete solids. It achieves 90\% success on a grasp-and-place task using the same planning, protocol, and verification stack.
Although a scripted policy slightly outperforms our method on the easiest task, its success rate falls to 10\% on the hardest, compared with 60\% for our method.
\end{abstract}

\section{Introduction}
\label{sec:intro}

\begin{figure}[tb]
  \centering
  \includegraphics[width=\textwidth]{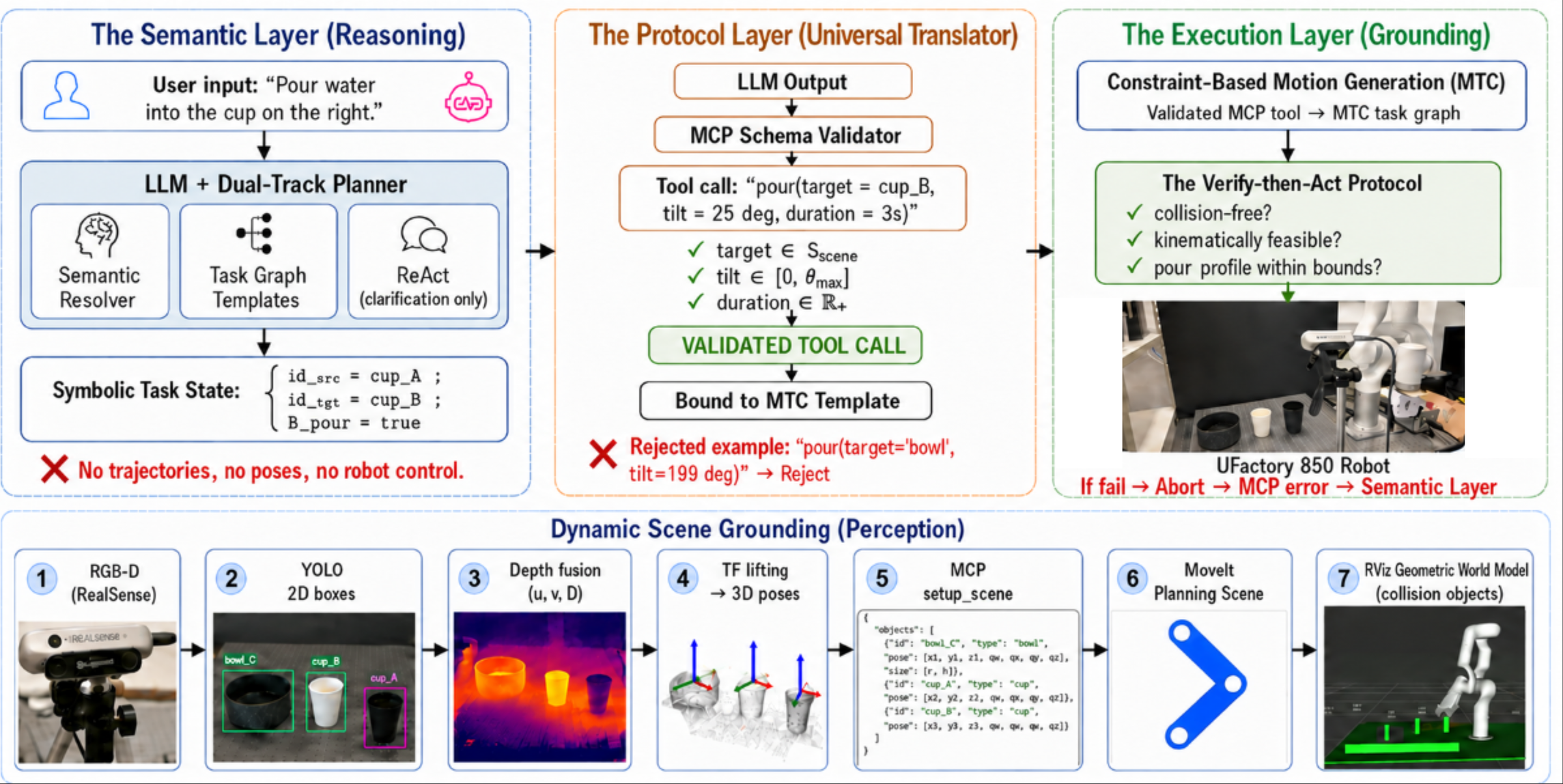}
  \caption{ Overview of the proposed framework. The Semantic Layer interprets the user instruction and produces a symbolic task state without generating poses or trajectories. The Protocol Layer validates MCP tool calls against a typed schema and binds valid calls to MTC templates. The Execution Layer follows a verify-then-act protocol, executing only motions that satisfy collision, kinematic, and task-specific constraints. The perception pipeline (bottom) grounds RGB-D observations into an explicit 3D MoveIt planning scene for collision-aware verification and execution.
  }
  \label{fig:teaser}
  \vspace{-0.5cm}
\end{figure}

Robotic assistance in everyday environments requires robots to interpret ambiguous natural-language instructions and execute them safely and reliably in cluttered, contact-rich, and uncertain settings~\cite{monwilliams2025embodied,brohan2023can,lin2023attention}. Consider the request in \cref{fig:teaser}: \emph{``Pour the water into the cup on the right.''} Executing this instruction requires (i) grounding linguistic references in physical entities and spatial relations, (ii) translating underspecified intent into a concrete, multi-step action sequence, and (iii) ensuring kinematic feasibility and collision avoidance during safety-critical manipulation tasks such as pouring and cooking~\cite{kim2024survey,zhang2021explainable,zhang2024self}.
The third requirement is particularly important for trustworthy robot execution. A plan may be linguistically correct yet physically unsafe, making it unsuitable for real-world deployment. 


Recent work on language-guided manipulation has shown that LLMs can decompose natural-language instructions into symbolic plans or structured tool calls~\cite{liang2023codeaspolicies,singh2022progprompt,song2023llmplanner,ahn2022saycan}. However, a persistent language-action gap remains: a plan may be semantically plausible yet physically infeasible under kinematic and collision constraints~\cite{jin2023robotgpt}.
One line of work pursues trustworthy robot execution through learned world models that predict the outcomes of candidate actions before execution. Although powerful, such models can hallucinate under partial observability and produce physically inconsistent predictions when they lack strong physics priors or explicit 3D structure, particularly in the safety-critical settings considered here~\cite{ai2025dynamics}.
We therefore take a complementary, explicit approach: rather than predicting action outcomes with a learned dynamics model, we verify each proposed action against a reconstructed, collision-aware representation of the current scene.
We formalize the planning-execution boundary as a typed contract, mapping language-level decisions into schema-validated tool calls through the Model Context Protocol (MCP)~\cite{anthropic2024mcp}.

Each validated call is then deterministically grounded in a MoveIt Task Constructor (MTC)~\cite{goerner2019mtc} pipeline, which checks kinematic feasibility and collisions before any motion is executed on the robot.

\subsubsection{Contributions.} This paper makes three contributions:

\begin{enumerate}
\item \textbf{A typed planning contract that bridges the language-action gap.} 
Our core contribution is a validated interface between language reasoning and robot execution: language-level decisions are mapped to schema-validated tool calls, so malformed or unsupported commands are rejected before reaching the motion planner.
We implement this interface through MCP as a portable schema layer.
Because the interface exposes semantic object references and bounded skill-level parameters rather than poses or trajectories, it can extend across skills by adding a corresponding tool schema and motion template without modifying the planner or execution stack.

\item \textbf{Verify-then-act grounding with explicit pre-execution checks.} Each validated tool call is deterministically grounded in an MTC template, and every candidate motion is evaluated against an explicit, perception-grounded model of the current scene. Only motions that satisfy kinematic-feasibility and collision constraints are sent to the robot, providing an auditable safety gate between probabilistic language reasoning and physical execution.

\item \textbf{Evidence for structured interfaces and recovery.} We evaluate the system on real-robot pouring tasks involving liquids, granular media, discrete solids, and demonstrate transfer to a grasp-and-place task while retaining the same planning and verification stack. Comparisons with a scripted policy and an unstructured language-model baseline, together with ablations of safe-pose recovery and retry mechanisms, show that our proposed framework improves overall task execution reliability.

\end{enumerate}


\section{Related Work}
\label{sec:related}

\subsection{Language-Guided Planning and Verified Execution}

LLMs have increasingly been used as high-level robot planners that translate natural-language instructions into programs or structured action sequences. Code-as-Policies~\cite{liang2023codeaspolicies} composes predefined robot APIs into executable policies, while Plan-Seq-Learn~\cite{dalal2024plan} further combines language-level planning with motion planning and learned low-level control.
Recent work couples language reasoning more directly with Task and Motion Planning (TAMP), which integrates symbolic task reasoning with geometric feasibility. AutoTAMP~\cite{chen2024autotamp} translates language instructions into formal task representations for downstream TAMP and iteratively corrects invalid specifications. LLM$^3$~\cite{wang2024llm3} allows the LLM to propose symbolic actions and continuous parameters, using motion-planning failures for refinement, while LLM-GROP~\cite{zhang2025llm} incorporates visually grounded commonsense reasoning into TAMP for long-horizon rearrangement. MoveIt Task Constructor (MTC)~\cite{goerner2019mtc} provides a practical framework for constructing and validating multi-stage manipulation motions.

Our work differs primarily in how language reasoning is exposed to robot execution. 
The LLM is restricted to schema-validated tool calls containing semantic object references and bounded skill-level parameters; it does not directly specify robot poses, joint configurations, waypoints, or trajectories. Each accepted call is mapped to a predefined MTC template, with geometric motion generation and verification handled entirely by the execution layer.
This design provides an explicit and auditable boundary between probabilistic language reasoning and geometric robot execution.

\subsection{Perception and Scene Grounding}

Language-guided manipulation requires linguistic references to be grounded in the observed environment. Prior work combines language with object detection, segmentation, pose estimation, or spatial representations to support visually grounded manipulation~\cite{huang2023voxposer,jiang2023vima,wake2024gpt4v}. For collision-aware motion planning, these observations must additionally be represented in a consistent geometric frame.

In our system, RGB-D detections are transformed into the robot frame and inserted into the MoveIt planning scene as collision geometry. The LLM accesses this grounded scene through typed tools, providing a shared representation for semantic grounding and geometric verification while keeping perception and planning explicitly separated.

\subsection{Robotic Pouring and Material Transfer}

Robotic pouring remains challenging because liquid and granular dynamics are difficult to model and only partially observable. Prior work has explored reinforcement learning~\cite{babaians2022pournet}, model-based control~\cite{chen2019pouring}, audio and haptic feedback~\cite{liang2019audio}, hierarchical imitation learning~\cite{zhang2022one}, and vision-based liquid perception~\cite{lin2023pourit,zhu2023pourme}. These methods primarily address perception or low-level control once the transfer objective is specified.

Our work addresses the complementary problem of executing language-specified transfer tasks through a structured and verified reasoning--execution interface. Rather than introducing a new pouring controller or dynamics model, we use pouring as a challenging physical domain for evaluating semantic grounding, structured action sequencing, pre-execution verification, and recovery.

\section{Method}
\label{sec:method}

\subsection{System Overview}
We study language-guided pouring in domestic environments, where success requires both correct semantic grounding and physically feasible motion. An instruction such as ``pour the water into the cup on the right'' demands resolving referential and spatial ambiguity, yet the resulting trajectory must also respect kinematic limits and collision constraints. End-to-end vision-language-action models aim to unify these capabilities, but balancing open-ended semantic reasoning with reliable physical execution remains challenging~\cite{brohan2022rt1,zitkovich2023rt2,driess2023palme,monwilliams2025embodied}.

We therefore make the reasoning-execution boundary explicit and typed: the language module produces structured, schema-validated action specifications, while the control stack executes only those that satisfy predefined feasibility and safety constraints. As illustrated in \cref{fig:framework}, the architecture comprises three layers. The Semantic Layer (\cref{sec:semantic}) interprets user intent and outputs only semantic action requests, without specifying robot poses or trajectories. The Protocol Layer (\cref{sec:protocol}) accepts only tool calls that satisfy the predefined schema and argument constraints. The Execution Layer (\cref{sec:execution}) maps each validated call to a predefined motion-planning template, generates the corresponding robot motion, and executes it only after kinematic and collision verification against the current planning scene. This separation keeps semantic reasoning independent while placing an explicit verification gate before physical execution.

\subsection{The Semantic Layer}
\label{sec:semantic}
\begin{figure}[tb]
  \centering
  \includegraphics[width=0.88\textwidth]{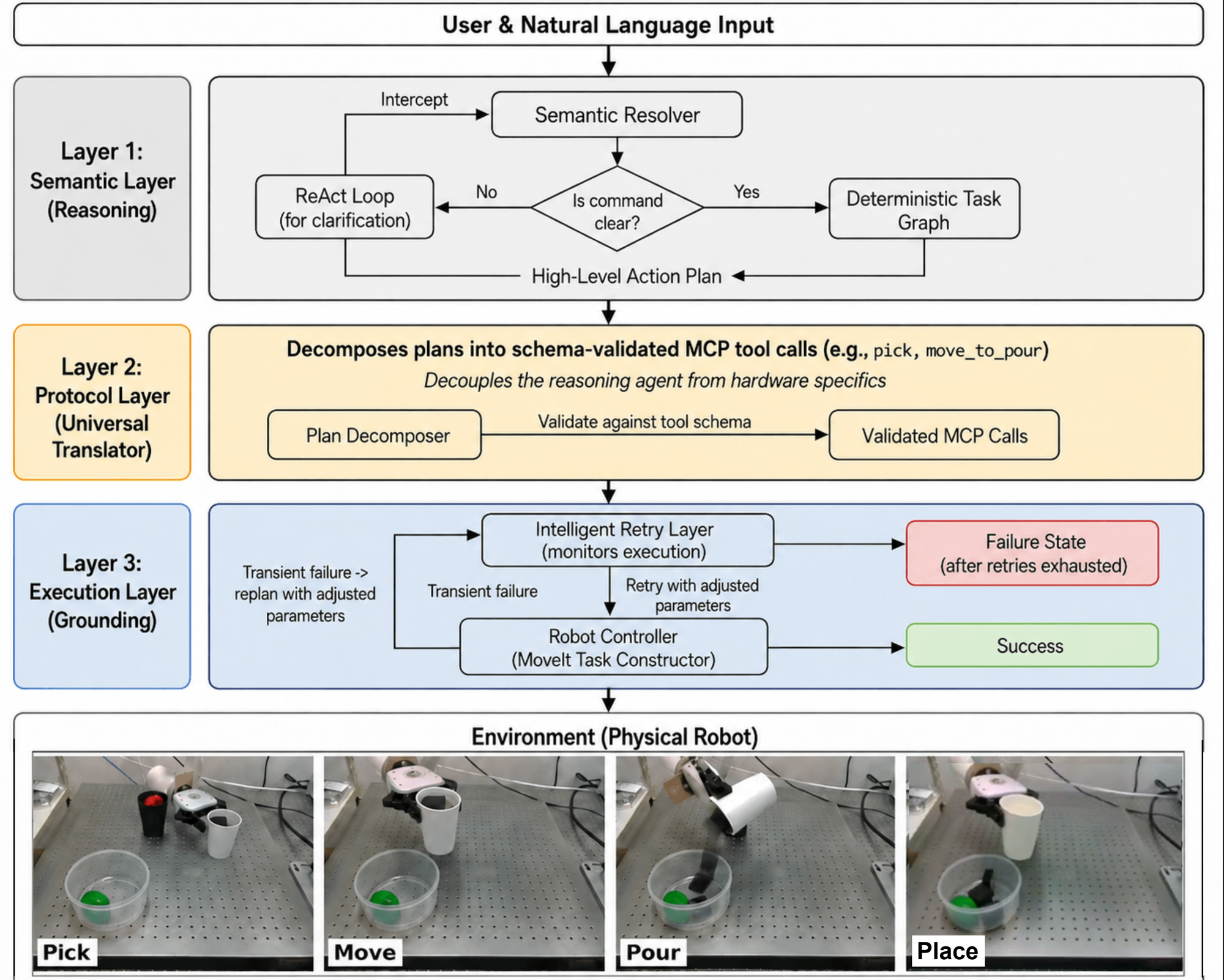}
  \caption{Control flow through the three layers, complementing the contract view of \cref{fig:teaser}. A semantic resolver routes each instruction either to a constrained ReAct loop that only queries information, or to a deterministic task graph. The protocol layer decomposes the resulting plan into schema-validated tool calls. The execution layer monitors execution and re-plans transient failures, committing to hardware only after verification. The bottom row shows the canonical pick, move, pour, and place sequence executing on the real robot.}
  \label{fig:framework}
    \vspace{-0.5cm}
\end{figure}

The Semantic Layer translates unstructured and potentially ambiguous instructions into a structured symbolic representation. Because unconstrained LLM outputs may contain invalid object references, unsupported actions, or out-of-range parameters, we organize this layer as a dual-track reasoning pipeline (\cref{fig:framework}).

\noindent\textbf{Language-model instantiation.}
We use Claude Sonnet 4 as the reasoning backbone of the Semantic Layer. The model operates on natural-language instructions together with a compact, perception-grounded representation of the current scene rather than directly processing raw visual observations. RGB-D inputs are handled separately by the perception stack, which detects objects, estimates their geometry, and registers them in the planning scene. The resulting structured context, including grounded object identifiers, semantic classes, and relevant robot state, is provided to the language model for high-level reasoning and tool selection. Image interpretation, pose estimation, collision checking, and trajectory generation remain delegated to the corresponding perception and execution modules.

\noindent\textbf{Semantic resolution and state abstraction.} Taking pouring as an example, the semantic resolver maps a natural-language instruction $U$ to a compact task-state tuple $S$ for downstream planning:
\begin{equation}
  S = \{\mathrm{id}_{\mathrm{src}},\ \mathrm{id}_{\mathrm{tgt}},\ B_{\mathrm{pour}}\},
  \label{eq:state}
\end{equation}
where $\mathrm{id}_{\mathrm{src}}$ denotes the source container, $\mathrm{id}_{\mathrm{tgt}}$ denotes the target container, and both are symbolic object identifiers grounded in the perception-injected planning scene. $B_{\mathrm{pour}}$ is a Boolean flag indicating whether the instruction requires material transfer. For the instruction of \cref{fig:teaser}, the resolver binds $\mathrm{id}_{\mathrm{src}}=\texttt{cup\_A}$ and $\mathrm{id}_{\mathrm{tgt}}=\texttt{cup\_B}$, and sets $B_{\mathrm{pour}}=\texttt{true}$. This abstraction retains only the task-relevant semantics, specifically which object transfers material to which target, while leaving poses, approach directions, and trajectories to the execution stack.

\noindent\textbf{Dual-track policy.} Based on instruction specificity, the planner follows one of two paths: a clarification track for ambiguous or underspecified requests and a deterministic execution track for well-specified commands. For ambiguous requests, the system invokes a constrained ReAct-style reasoning agent~\cite{yao2023react} that alternates between reasoning and information gathering. The agent may query perception or ask the user for clarification, but it cannot invoke actuation tools or issue motion commands.
The clarification track is used specifically to resolve referential ambiguity rather than to perform long-horizon task planning. For example, given the instruction ``pour the coffee into the cup'' when two cups are visible, the agent queries the scene, identifies two matching object identifiers, and asks the user to specify the intended target. Only after clarification does it bind $\mathrm{id}_{\mathrm{tgt}}$ and transfer control to the execution track. If the ambiguity remains unresolved, the task is halted rather than executed using a guessed target.
For well-specified commands, such as the instruction in \cref{fig:teaser}, the planner directly selects a pre-validated finite-state-machine template that defines the canonical action sequence (pick, move, pour, place). The LLM determines the task-level intent but does not generate the execution sequence itself. Instead, execution follows a predefined motion template that is geometrically verified before being sent to the robot.

\subsection{The Protocol Layer}
\label{sec:protocol}
Every action passes through a typed interface that enforces structural validity and maps each accepted call to a predefined motion-planning procedure. We implement this interface using the MCP, which exposes robot capabilities as a finite set of schema-defined tools. Each tool call is validated against its schema and scene-dependent preconditions before being passed to the execution layer~\cite{li2023apibank,crouse2023llmagents}.
Tool calls contain semantic object references and bounded skill-level parameters, while robot poses, joint configurations, collision checking, and trajectory generation remain encapsulated within the corresponding motion-planning templates.
This interface therefore provides an explicit barrier between probabilistic language reasoning and safety-critical robot execution.

\noindent\textbf{Schema-based action definition.} Each capability is exposed as a tool defined by a rigorous JSON schema. Let $S_{\mathrm{scene}}$ denote the set of object identifiers currently registered in the perception-grounded planning scene. A typical manipulation tool is defined over a typed domain,
\begin{equation}
  \mathrm{Tool}_{\mathrm{pour}} : \bigl(\mathrm{id}_{\mathrm{tgt}} \in S_{\mathrm{scene}},\ \theta_{\mathrm{tilt}} \in [0, \theta_{\max}],\ \tau_{\mathrm{duration}} \in \mathbb{R}^{+}\bigr).
  \label{eq:tool}
\end{equation}
At runtime, the MCP server applies three stages of validation. First, schema validation checks required fields and data types. Second, range validation ensures that physical parameters, such as tilt limits and action duration, remain within predefined admissible bounds. Third, referential validation verifies that semantic references, such as $\mathrm{id}{\mathrm{tgt}}$, correspond to objects present in $S{\mathrm{scene}}$. Calls that fail any of these checks are rejected before reaching the motion planner, preventing invalid commands from propagating to the robot.

\noindent\textbf{Atomic-to-composite binding.} The Semantic Layer represents each tool call as an atomic action, while physical execution unfolds as a multi-stage motion program. Each tool is therefore associated with a predefined MTC template (\cref{tab:binding}). From the planner's perspective, the action is atomic; from the executor's perspective, it is decomposed into a sequence of motion stages. A validated call is passed to the corresponding task builder, which instantiates the template and searches for a kinematically feasible, collision-free solution. For example, a \texttt{pick} call expands into a sequence of approach, grasp, lift, and retreat stages that are verified before execution.

\begin{table}[tb]
  \caption{Mapping from MCP tools to MTC task graphs. Each atomic tool call at the planner level is bound to a composite pipeline of MTC stages at the executor level.}
  \label{tab:binding}
  \centering
  \begin{tabular}{@{}l p{0.63\linewidth}@{}}
    \toprule
    MCP tool interface (atomic) & Underlying MTC stage pipeline (composite)\\
    \midrule
    \texttt{setup\_scene}  & Perception injection $\rightarrow$ collision-object spawning\\
    \texttt{pick(object)}  & Grasp generation $\rightarrow$ IK filtering $\rightarrow$ approach $\rightarrow$ close $\rightarrow$ lift\\
    \texttt{move\_to\_pour} & Cartesian planning $\rightarrow$ tilt-constraint check $\rightarrow$ velocity scaling $\rightarrow$ execute\\

    \texttt{place(target)} & Place-pose generation $\rightarrow$ approach $\rightarrow$ open $\rightarrow$ retreat\\
  \bottomrule
  \end{tabular}
    \vspace{-0.5cm}
\end{table}

\noindent\textbf{Auditability and platform agnosticism.} The Protocol Layer also standardizes runtime behavior: every invocation is logged with its semantic parameters, and failures are normalized into structured responses such as \texttt{PlanningFailed} or \texttt{TargetUnreachable}, so the agent handles errors by retrying or querying the user, and never parses a driver exception. Because the interface hides the controller implementation, the same plan binds to different backends; we exploit this to bind tools first to a kinematic verification environment for pre-flight checking, then to the physical UFactory 850 driver.

\subsection{The Execution Layer}
\label{sec:execution}
The Execution Layer translates validated tool calls into safe, executable motions in three stages: it grounds the scene from perception, generates motion under explicit constraints, and verifies each candidate plan before committing it to hardware.

\begin{figure}[tb]
  \centering
  \includegraphics[width=0.95\textwidth]{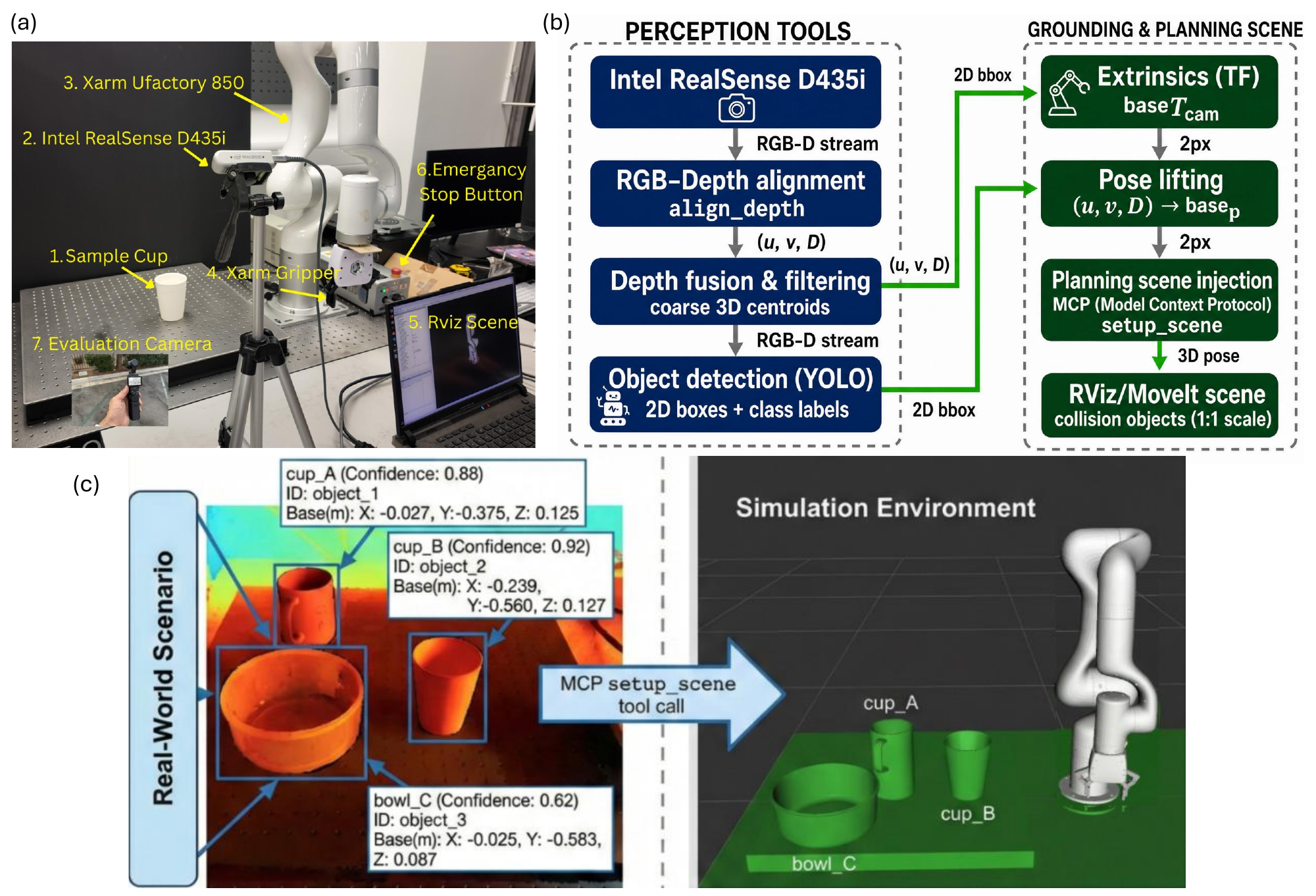}
  \caption{ (a) Real-world experimental setup using a UFactory 850 robot equipped with a two-finger gripper and an Intel RealSense D435i for perception. A separate camera records the experiments for evaluation. (b) Modular perception tools process RGB-D observations, transform detected objects into the robot base frame using calibrated extrinsics, and insert them into the planning scene through the \texttt{setup\_scene} tool. (c) Detected objects and their estimated base-frame poses are shown on the colorized depth image (left), with the corresponding collision geometry in the planning scene (right). }
  \label{fig:setup}
    \vspace{-0.5cm}
\end{figure}

\noindent\textbf{Dynamic scene grounding.} Symbolic references such as \texttt{cup\_A} must be grounded in geometry before the planner can reason about them. 
The experimental workspace shown in \cref{fig:setup}(a) comprises a UFactory 850 robot and an Intel RealSense D435i in an eye-to-hand configuration.
As shown in \cref{fig:setup}(b), RGB streams are processed by a YOLO-based detector~\cite{jocher2023yolov8} under class-specific confidence thresholds, and detections are lifted into three dimensions by fusing aligned depth, median-filtered within each bounding box to suppress noise. The recovered points are transformed into the robot base frame via extrinsics from a $5\times7$ ChArUco board, accurate to sub-centimeter translation and one to two degrees. Rather than feeding these estimates to the controller, we inject them through the \texttt{setup\_scene} tool, which instantiates collision geometries at unit scale using predefined cylindrical proxies rather than online-reconstructed meshes. 
The resulting perception-grounded planning scene (\cref{fig:setup}(c)) provides an explicit geometric representation of the current workspace for collision checking and kinematic reachability analysis.

\noindent\textbf{Constraint-based motion generation.} Rather than optimizing one trajectory end to end, MTC factorizes each skill into stages with explicit feasibility checks. \emph{Generators} seed candidates: a grasp-pose generator reads the object pose from the planning scene and samples grasp hypotheses, pruned by inverse-kinematics (IK) reachability. \emph{Propagators} solve local subproblems such as a safe-height approach, a descent, or a post-closure lift, each admitting its own velocity, acceleration, and orientation constraints. \emph{Connectors} bridge disjoint states and backtrack to the next candidate when no feasible bridge exists, which keeps failure modes attributable to a stage.

For material transfer, we extend this pipeline with a conservative staged pouring profile~\cite{chen2019pouring,babaians2022pournet}. We do not model the detailed dynamics of the transferred material. Instead, the end-effector executes a sequence of tilt-and-hold stages subject to predefined bounds on tilt angle, angular velocity, and dwell time, with each stage additionally checked for kinematic feasibility and collisions. These constraints limit abrupt motions during pouring while keeping the execution policy simple and inspectable. The verify-then-act mechanism therefore validates whether the commanded pouring motion is executable under the current kinematic and collision constraints, rather than predicting the resulting material dynamics.

\noindent\textbf{Verify-then-act protocol.} To reduce the risk of executing automatically generated plans in cluttered scenes, we enforce a strict kinematic and collision-verification step prior to hardware actuation~\cite{jin2023robotgpt,khan2024safetask}. Once a candidate plan is produced, it is replayed against the reconstructed planning scene to verify that the full motion, from approach through retreat, remains continuous, reachable, and collision-free with respect to the modeled geometry. Only trajectories that pass these checks are serialized and transmitted to the UFactory 850 controller. If verification detects a collision, singularity, or discontinuity, execution is aborted and a structured failure is returned to the Semantic Layer. In addition, a predefined collision-checked safe pose is inserted between high-risk stages, for example after grasping and before pouring, to reduce error accumulation and support recovery. This verify-then-act loop ensures that only motions validated for continuity, reachability, and collision avoidance in the reconstructed scene reach the hardware, preventing hallucinated or physically infeasible commands from being executed.

\section{Experiments}
\label{sec:exp}

\subsection{Experimental Setup}
All experiments are conducted on the UFactory 850 platform shown in \cref{fig:setup}(a), using the perception, protocol, and execution stack described in \cref{sec:method}.

We evaluate four end-to-end pouring tasks of increasing difficulty (T1--T4), ranging from single-source pouring in a minimally cluttered scene to multi-source, multi-material, and cluttered scenarios. Each task is evaluated over ten independent physical trials. A trial comprises the complete pick, move-to-pour, pour, and place sequence and is counted as successful only if the material reaches the target container and the source container is returned. Because $N=10$, each trial corresponds to ten percentage points; we therefore interpret small per-task differences cautiously and focus on consistent trends across tasks.

We compare our method with two baselines that share the same robot, perception pipeline, and motion-planning backend. The \emph{FSM/BT} baseline executes a hand-crafted sequence of the same motion primitives, while 
the \emph{LLM without task graph} retains the same language model, tool schemas, schema validation, safe-pose recovery, retry mechanism, motion-planning backend, and verify-then-act checks as our full method; only canonical task-graph sequencing is removed.
End-to-end task success is the primary metric. Perception uncertainty is characterized separately through pose-grounding error in \cref{fig:poseerr}.

\subsection{Perception and Scene Grounding}
\begin{figure}[tb]
  \centering
  \begin{subfigure}{0.48\linewidth}
    \centering
    \includegraphics[width=\linewidth]{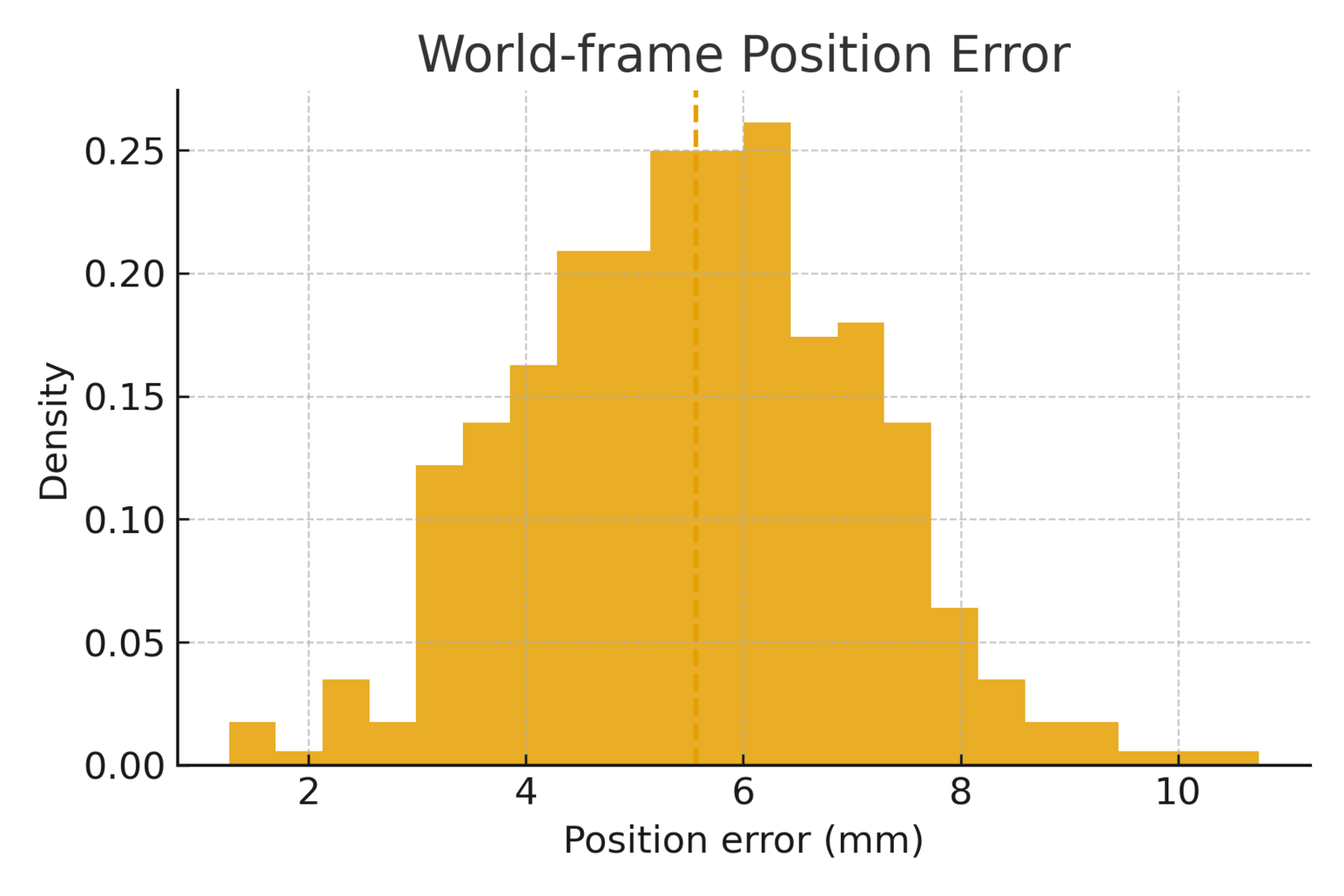}
    \caption{World-frame position error}
    \label{fig:poseerr-pos}
  \end{subfigure}
  \hfill
  \begin{subfigure}{0.48\linewidth}
    \centering
    \includegraphics[width=\linewidth]{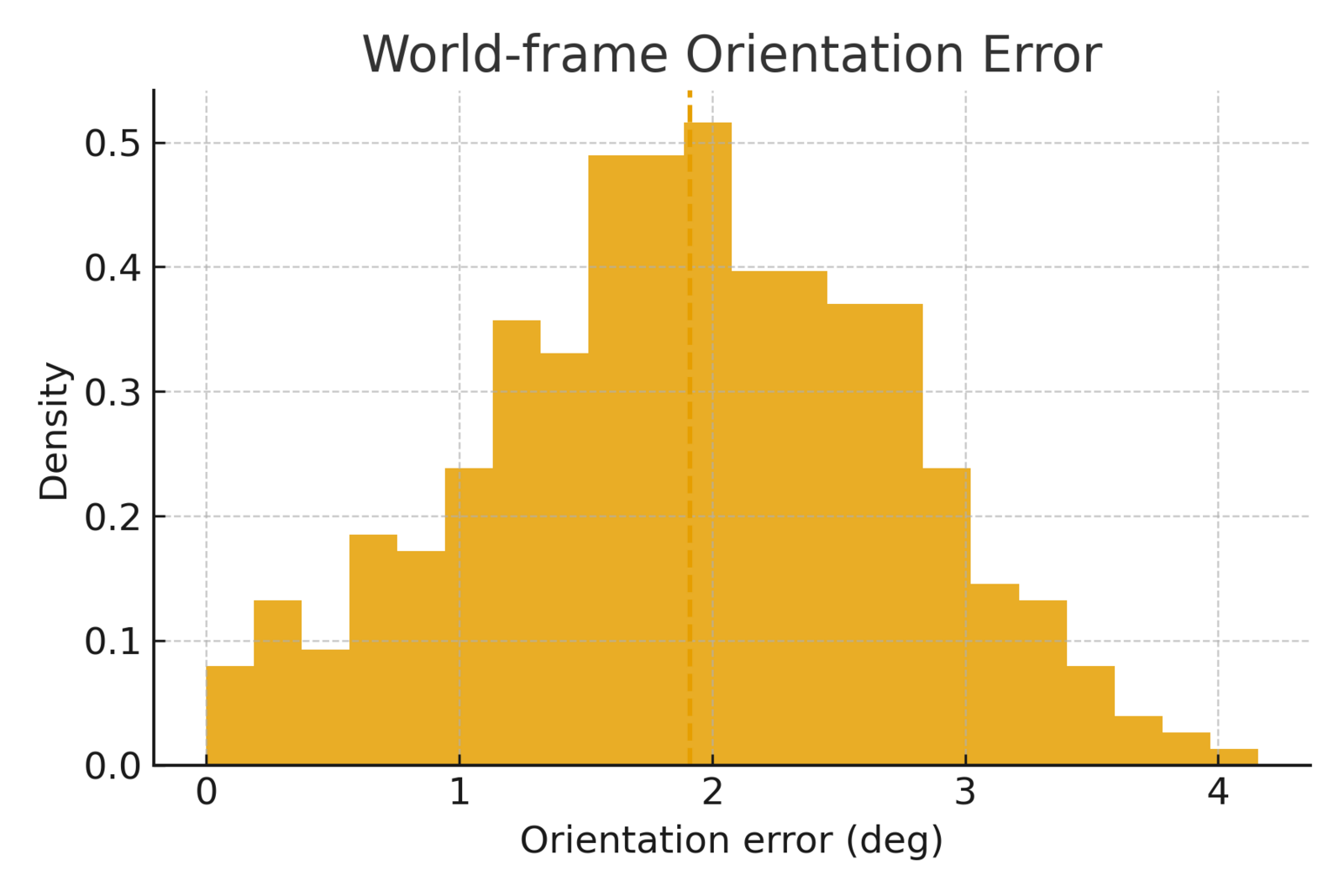}
    \caption{World-frame orientation error}
    \label{fig:poseerr-ori}
  \end{subfigure}
  \caption{World-frame object pose grounding errors across all scenarios. Both distributions exhibit non-negligible tails, motivating protocol-level verification prior to execution.}
  \label{fig:poseerr}
      \vspace{-0.2cm}
\end{figure}

Grounding accuracy is evaluated by comparing estimated object poses with manually verified references across all scenarios. Position error is measured as the Euclidean distance between object centroids, and orientation error as the angular deviation between object frames (\cref{fig:poseerr}). Most estimates are sufficiently accurate for collision-aware planning and stable pouring, although larger errors occur under clutter and partial occlusion. Perception can introduce both semantic and geometric errors: visually similar containers may be associated with the wrong language reference, while pose-estimation errors can create discrepancies between the reconstructed scene and the physical workspace. The Protocol Layer rejects missing or inconsistent object references before motion planning, whereas the Execution Layer verifies candidate motions against the reconstructed planning scene. Residual pose uncertainty nevertheless remains a source of model--world mismatch, motivating explicit geometric verification and conservative execution.

\subsection{Task Performance}
\begin{figure}[tb]
  \centering
  \includegraphics[width=0.93\textwidth]{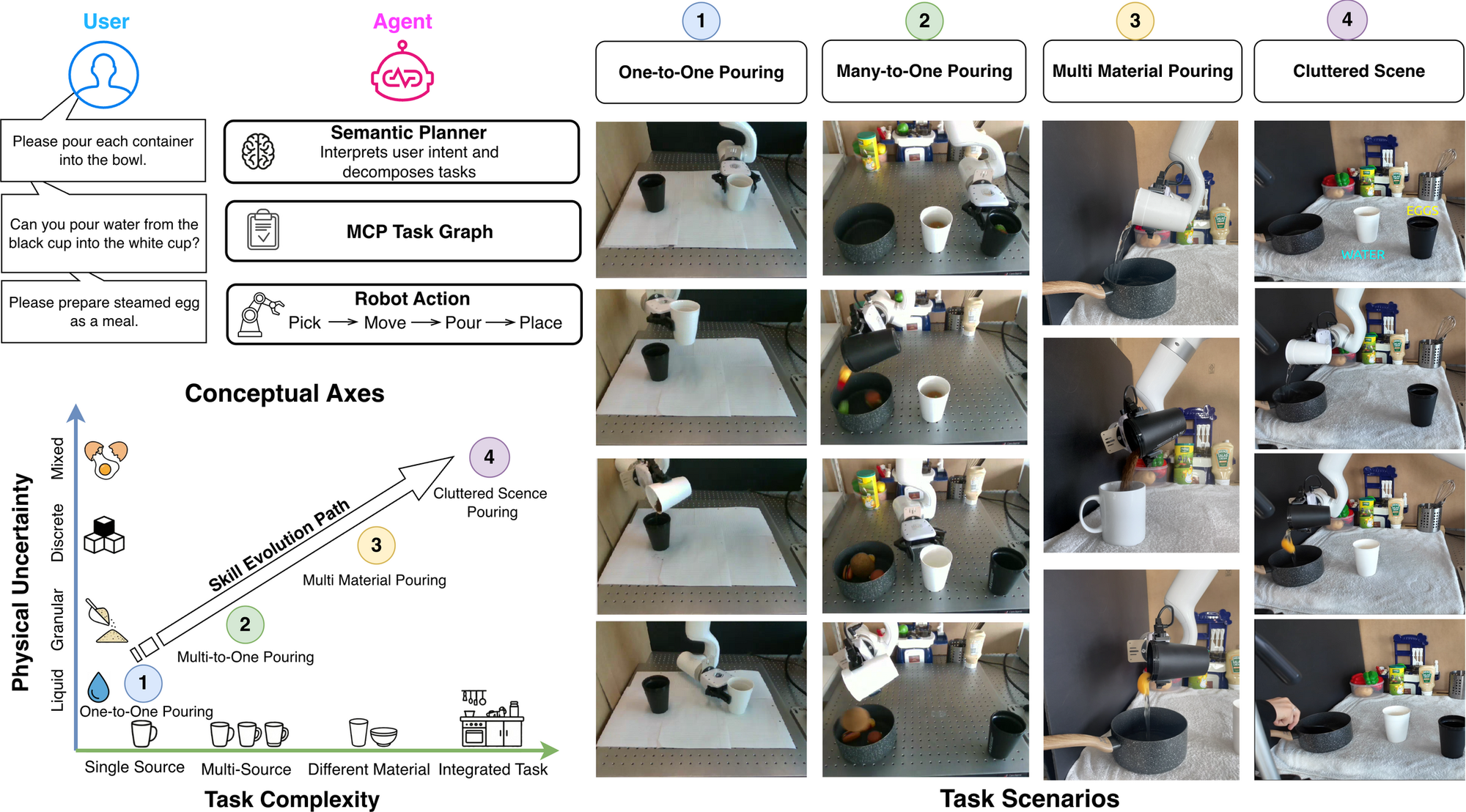}
\caption{Real-robot task suite. Left: two dimensions of increasing difficulty, task complexity (single source, multiple sources, different materials, and integrated task) and material uncertainty (liquid, granular, discrete, and mixed). Right: representative executions of the four evaluated scenarios, T1-T4.}
  \label{fig:tasks}
      \vspace{-0.3cm}
\end{figure}

We evaluate end-to-end performance on four real-robot tasks that progressively introduce longer horizons, denser clutter, and higher execution uncertainty (\cref{fig:tasks}). \Cref{tab:main} reports success against both baselines.

\begin{table}[tb]
\caption{Real-robot task performance. Each entry reports end-to-end success over 10 independent physical trials, with each trial corresponding to 10 percentage points. Per-task differences should therefore be interpreted cautiously. \emph{FSM/BT} denotes a hand-crafted scripted policy, while \emph{LLM w/o task graph} uses the same tool set without canonical action sequencing. Best result in each row is shown in bold.}
  \label{tab:main}
  \centering
  \begin{tabular}{@{}lccc@{}}
    \toprule
    Task & Ours & FSM/BT & LLM w/o task graph\\
    \midrule
    T1: Single Pour        & 80\%          & \textbf{90\%} & 60\%\\
    T2: Multi-Source Pour  & \textbf{70\%} & 40\%          & 30\%\\
    T3: Multi-Material Pour& \textbf{80\%} & 60\%          & 60\%\\
    T4: Complex Scene      & \textbf{60\%} & 10\%          & 40\%\\
    \midrule
    Mean                   & \textbf{72.5\%} & 50.0\%      & 47.5\%\\
  \bottomrule
  \end{tabular}
    \vspace{-0.5cm}
\end{table}

The ranking is not uniform, and the exception is informative. On T1, where the target is unambiguous and the scene is sparse, the scripted FSM/BT policy achieves 90\% success compared with our 80\%. In this simple setting, a hand-tuned sequence incurs no reasoning overhead, and the difference corresponds to a single trial. This advantage disappears on harder tasks: the scripted policy drops to 40\% on T2 and 10\% on T4, compared with 70\% and 60\% for our method, respectively. These tasks require adapting to distractors, rebinding references, and tighter geometric constraints that a fixed sequence cannot accommodate. The unstructured LLM exhibits the opposite weakness, performing below our method on every task because the absence of canonical sequencing allows grounding and ordering errors to propagate into invalid intermediate states. Averaged across all tasks, our method achieves 72.5\% success, compared with 50.0\% for the scripted policy and 47.5\% for the unstructured LLM, with the largest gains occurring in the more constrained scenes.

\subsection{Generalization Beyond Pouring}
\begin{figure}[tb]
  \centering
  \includegraphics[width=\textwidth]{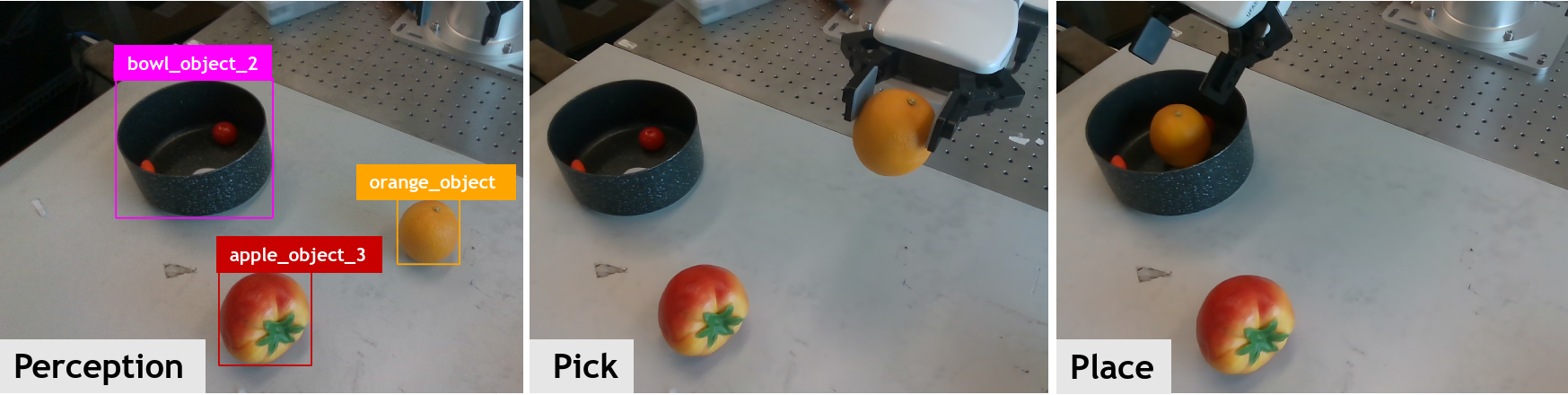}
\caption{Generalization to a fruit grasp-and-place task. Only the skill-specific tool schema and MTC template are changed. Perception grounds the fruit and bowl in the planning scene, while validated pick-and-place stages transfer the fruit to the bowl. The planner, protocol layer, and verification stack remain unchanged.}
  \label{fig:general}
    \vspace{-0.5cm}
\end{figure}

Since our contribution lies in the validated reasoning-execution interface rather than in pouring itself, we also test whether the same pipeline transfers to a different manipulation skill. We evaluate a fruit grasp-and-place task in which the robot moves an apple or orange into a bowl (\cref{fig:general}). Transfer requires only a task-specific tool schema and corresponding MTC template; the planner, protocol layer, and verify-then-act mechanism remain unchanged. Since this task differs from the pouring suite in both objects and manipulation behavior, its results are not directly comparable to \cref{tab:main}. Instead, the experiment tests whether the interface can be reused across skills. 

Over ten trials, the system achieves 90\% end-to-end success. The single failure was caused by a pose-estimation error near the camera field-of-view boundary, which resulted in a collision during grasp approach. Although this experiment is not intended as a broad multi-skill benchmark, it provides evidence that the framework can transfer to a new manipulation behavior by adding only a new tool schema and MTC template while retaining the same validation and verify-then-act mechanisms.

\subsection{Multi-Material Robustness}
T3 directly tests robustness to material variation by transferring water, coffee beans, and discrete solids such as eggs, which exhibit substantially different flow characteristics and failure modes. The method achieves 80\% success on T3 (\cref{tab:main}), matching its performance on the simpler single-source T1. This suggests that performance is relatively insensitive to material type, since pouring behavior is governed by parameterized tilt limits and execution-time verification rather than material-specific tuning. The remaining failures are also not strongly material dependent. Liquids benefit from conservative velocity scaling and stabilization at the pour pose, whereas granular and discrete materials are more tolerant of tilt timing but more sensitive to grasp stability; in these cases, failures are more often associated with grasping than with the pouring motion itself.

Across all three material classes, we use the same staged pouring profile and task-level interface without introducing material-specific planning logic. T3 achieves 80\% end-to-end success despite the distinct transfer characteristics of liquids, granular media, and discrete solids. This result suggests that the same perception-planning-execution abstraction can generalize across substantially different material behaviors without task-specific retuning.

\subsection{Ablation Studies}
\Cref{tab:main} compares alternative reasoning structures under shared perception and motion-planning backends; here, we further examine which protocol-level mechanisms contribute to end-to-end robustness. The scripted policy provides a lower bound on reasoning capability: without language-based adaptation, its success rate falls to 10\% on T4, compared with 60\% for our method. At the other extreme, the same language model without the task graph retains flexible reasoning but performs worse than our method on every task. Together, these comparisons suggest that language reasoning alone is insufficient; structured action sequencing and execution-time recovery are key contributors to reliable end-to-end performance.

\Cref{tab:ablation} isolates the contribution of the two protocol-level recovery mechanisms by removing them individually while retaining schema validation and verify-then-act. Both have a larger effect on performance than the task graph: removing the safe-pose reset reduces average success by 32.5 percentage points, while disabling retry reduces it by 47.5 points, compared with a 25.0-point drop when the task graph is removed. Retry has the largest effect because many failures on this platform are transient. For example, an inverse-kinematics failure at a grasp pose or a planning failure caused by insufficient clearance can often be recovered by replanning with an alternative feasible candidate. Without retry, such failures terminate the trial. The safe-pose reset provides a complementary recovery mechanism by returning the arm to a predefined, collision-checked configuration between stages, reducing the likelihood of entering configurations from which subsequent motions cannot be planned.

These results indicate that robustness arises from both structured sequencing and protocol-level recovery, with recovery showing the larger effect in our experiments. Schema validation provides a front-end safeguard by rejecting malformed calls before they reach the motion planner, while recovery mechanisms address transient failures during physical execution. Canonical sequencing also contributes to overall reliability, although its measured effect is smaller. Given only ten trials per condition, individual per-task differences should be interpreted cautiously. We therefore limit our conclusions to the overall performance trends across tasks.

\begin{table}[tb]
  \caption{Ablations of our own system, each over 10 independent physical trials. Removing the safe-pose reset or the retry mechanism leaves schema validation and verify-then-act intact, so these isolate the protocol-level recovery mechanisms.}
  \label{tab:ablation}
  \centering
  \begin{tabular}{@{}lccc@{}}
    \toprule
    Task & Ours & w/o safe-pose & w/o retry\\
    \midrule
    T1: Single Pour        & \textbf{80\%} & 50\% & 20\%\\
    T2: Multi-Source Pour  & \textbf{70\%} & 20\% & 40\%\\
    T3: Multi-Material Pour& \textbf{80\%} & 70\% & 10\%\\
    T4: Complex Scene      & \textbf{60\%} & 20\% & 30\%\\
    \midrule
    Mean                   & \textbf{72.5\%} & 40.0\% & 25.0\%\\
  \bottomrule
  \end{tabular}
  \vspace{-0.6cm}
\end{table}

\section{Conclusions and Future Work}
\label{sec:conclusion}
We presented a typed reasoning-execution interface for language-guided manipulation that maps language-level intents into validated tool calls and predefined motion-planning templates. Rather than predicting physical outcomes with a learned dynamics model, the system verifies each candidate motion against a perception-grounded planning scene before execution, providing an explicit and auditable gate between probabilistic language reasoning and robot motion.
Across four real-robot pouring tasks, our method achieves 72.5\% mean end-to-end success, compared with 50.0\% for the scripted FSM/BT baseline and 47.5\% for the unstructured LLM baseline. Transfer to a grasp-and-place task reaches 90\% success while retaining the same planning, protocol, and verification stack. Ablations further show that both structured sequencing and protocol-level recovery contribute to overall reliability, with retry producing the largest measured effect in our experiments.

The current approach remains limited by the accuracy of the reconstructed planning scene, does not model material dynamics, and lacks closed-loop feedback during execution. Future work will incorporate visual, tactile, and audio feedback for closed-loop manipulation and material transfer, and evaluate robustness under greater occlusion, dynamic scene changes, and denser multi-object clutter.




\bibliographystyle{splncs04}
\bibliography{main}

@article{kim2024survey,
  author = {Kim, Yeseung and Kim, Dohyun and Choi, Jieun and Park, Jisang and Oh, Nayoung and Park, Daehyung},
  title = {A Survey on Integration of Large Language Models with Intelligent Robots},
  journal = {Intelligent Service Robotics},
  volume = {17},
  number = {5},
  pages = {1091--1107},
  year = {2024}
}

@article{zhang2021explainable,
  title={Explainable hierarchical imitation learning for robotic drink pouring},
  author={Zhang, Dandan and Li, Qiang and Zheng, Yu and Wei, Lei and Zhang, Dongsheng and Zhang, Zhengyou},
  journal={IEEE Transactions on Automation Science and Engineering},
  volume={19},
  number={4},
  pages={3871--3887},
  year={2021},
  publisher={IEEE}
}

@misc{ahn2022saycan,
  author = {Ahn, Michael and Brohan, Anthony and Brown, Noah and Chebotar, Yevgen and Cortes, Omar and David, Byron and Finn, Chelsea and others},
  title = {Do As {I} Can, Not As {I} Say: Grounding Language in Robotic Affordances},
  howpublished = {arXiv preprint arXiv:2204.01691},
  year = {2022}
}

@inproceedings{zhang2024self,
  title={Self-supervised bayesian visual imitation learning applied to robotic pouring},
  author={Zhang, Dan-Dan and Zheng, Yu and Fan, Wen and Lepora, Nathan and Zhang, Zhengyou},
  booktitle={2024 IEEE International Conference on Industrial Technology (ICIT)},
  pages={1--7},
  year={2024},
  organization={IEEE}
}

@article{zhang2022one,
  title={One-shot domain-adaptive imitation learning via progressive learning applied to robotic pouring},
  author={Zhang, Dandan and Fan, Wen and Lloyd, John and Yang, Chenguang and Lepora, Nathan F},
  journal={IEEE Transactions on Automation Science and Engineering},
  volume={21},
  number={1},
  pages={541--554},
  year={2022},
  publisher={IEEE}
}

@misc{anthropic2024mcp,
  author = {{Anthropic}},
  title = {Introducing the {M}odel {C}ontext {P}rotocol},
  howpublished = {\url{https://www.anthropic.com/news/model-context-protocol}},
  note = {Accessed 15 Aug 2026},
  year = {2024}
}

@inproceedings{babaians2022pournet,
  author = {Babaians, Edwin and Sharma, Tapan and Karimi, Mojtaba and Sharifzadeh, Sahand and Steinbach, Eckehard},
  title = {{PourNet}: Robust Robotic Pouring Through Curriculum and Curiosity-based Reinforcement Learning},
  booktitle = {IEEE/RSJ International Conference on Intelligent Robots and Systems (IROS)},
  pages = {9332--9339},
  year = {2022}
}

@misc{brohan2022rt1,
  author = {Brohan, Anthony and Brown, Noah and Carbajal, Justice and Chebotar, Yevgen and Dabis, Joseph and Finn, Chelsea and others},
  title = {{RT-1}: Robotics Transformer for Real-World Control at Scale},
  howpublished = {arXiv preprint arXiv:2212.06817},
  year = {2022}
}

@inproceedings{chen2019pouring,
  author = {Chen, Tianze and Huang, Yongqiang and Sun, Yu},
  title = {Accurate Pouring using Model Predictive Control Enabled by Recurrent Neural Network},
  booktitle = {IEEE/RSJ International Conference on Intelligent Robots and Systems (IROS)},
  pages = {7688--7694},
  year = {2019}
}

@misc{crouse2023llmagents,
  author = {Crouse, Maxwell and Abdelaziz, Ibrahim and Astudillo, Ramon and Basu, Kinjal and Dan, Soham and Kumaravel, Sadhana and Fokoue, Achille and Kapanipathi, Pavan and Roukos, Salim and Lastras, Luis},
  title = {Formally Specifying the High-Level Behavior of {LLM}-Based Agents},
  howpublished = {arXiv preprint arXiv:2310.08535},
  year = {2023}
}

@inproceedings{driess2023palme,
  author = {Driess, Danny and Xia, Fei and Sajjadi, Mehdi S.~M. and Lynch, Corey and Chowdhery, Aakanksha and Ichter, Brian and others},
  title = {{PaLM-E}: An Embodied Multimodal Language Model},
  booktitle = {International Conference on Machine Learning (ICML)},
  year = {2023}
}

@inproceedings{huang2023voxposer,
  author = {Huang, Wenlong and Wang, Chen and Zhang, Ruohan and Li, Yunzhu and Wu, Jiajun and Fei-Fei, Li},
  title = {{VoxPoser}: Composable {3D} Value Maps for Robotic Manipulation with Language Models},
  booktitle = {Conference on Robot Learning (CoRL)},
  year = {2023}
}

@inproceedings{jiang2023vima,
  author = {Jiang, Yunfan and Gupta, Agrim and Zhang, Zichen and Wang, Guanzhi and Dou, Yongqiang and Chen, Yanjun and Fei-Fei, Li and Anandkumar, Anima and Zhu, Yuke and Fan, Linxi},
  title = {{VIMA}: Robot Manipulation with Multimodal Prompts},
  booktitle = {International Conference on Machine Learning (ICML)},
  year = {2023}
}

@misc{jin2023robotgpt,
  author = {Jin, Yixiang and Li, Dingzhe and A, Yong and Shi, Jun and Hao, Peng and Sun, Fuchun and Zhang, Jianwei and Fang, Bin},
  title = {{RobotGPT}: Robot Manipulation Learning from {ChatGPT}},
  howpublished = {arXiv preprint arXiv:2312.01421},
  year = {2023}
}

@misc{khan2024safetask,
  author = {Khan, Azal Ahmad and Andrev, Michael and Murtaza, Muhammad Ali and Aguilera, Sergio and Zhang, Rui and Ding, Jie and Hutchinson, Seth and Anwar, Ali},
  title = {Safety Aware Task Planning via Large Language Models in Robotics},
  howpublished = {arXiv preprint arXiv:2503.15707},
  year = {2025}
}

@inproceedings{li2023apibank,
  author = {Li, Minghao and Zhao, Yingxiu and Yu, Bowen and Song, Feifan and Li, Hangyu and Yu, Haiyang and Li, Zhoujun and Huang, Fei and Li, Yongbin},
  title = {{API-Bank}: A Comprehensive Benchmark for Tool-Augmented {LLM}s},
  booktitle = {Conference on Empirical Methods in Natural Language Processing (EMNLP)},
  pages = {3102--3116},
  year = {2023}
}

@inproceedings{liang2019audio,
  author = {Liang, Hongzhuo and Li, Shuang and Ma, Xiaojian and Hendrich, Norman and Gerkmann, Timo and Sun, Fuchun and Zhang, Jianwei},
  title = {Making Sense of Audio Vibration for Liquid Height Estimation in Robotic Pouring},
  booktitle = {IEEE/RSJ International Conference on Intelligent Robots and Systems (IROS)},
  pages = {5333--5339},
  year = {2019}
}

@inproceedings{liang2023codeaspolicies,
  author = {Liang, Jacky and Huang, Wenlong and Xia, Fei and Xu, Peng and Hausman, Karol and Ichter, Brian and Florence, Pete and Zeng, Andy},
  title = {Code as Policies: Language Model Programs for Embodied Control},
  booktitle = {IEEE International Conference on Robotics and Automation (ICRA)},
  pages = {9493--9500},
  year = {2023}
}

@misc{lin2023pourit,
  author = {Lin, Haitao and Fu, Yanwei and Xue, Xiangyang},
  title = {{PourIt!}: Weakly-Supervised Liquid Perception from a Single Image for Visual Closed-Loop Robotic Pouring},
  howpublished = {arXiv preprint arXiv:2307.11299},
  year = {2023}
}

@article{monwilliams2025embodied,
  author = {Mon-Williams, Ruaridh and Li, Gen and Long, Ran and Du, Wenqian and Lucas, Christopher G.},
  title = {Embodied Large Language Models Enable Robots to Complete Complex Tasks in Unpredictable Environments},
  journal = {Nature Machine Intelligence},
  volume = {7},
  number = {4},
  pages = {592--601},
  year = {2025}
}

@misc{singh2022progprompt,
  author = {Singh, Ishika and Blukis, Valts and Mousavian, Arsalan and Goyal, Ankit and Xu, Danfei and Tremblay, Jonathan and Fox, Dieter and Thomason, Jesse and Garg, Animesh},
  title = {{ProgPrompt}: Generating Situated Robot Task Plans using Large Language Models},
  howpublished = {arXiv preprint arXiv:2209.11302},
  year = {2022}
}

@misc{song2023llmplanner,
  author = {Song, Chan Hee and Wu, Jiaman and Washington, Clayton and Sadler, Brian M. and Chao, Wei-Lun and Su, Yu},
  title = {{LLM-Planner}: Few-Shot Grounded Planning for Embodied Agents with Large Language Models},
  howpublished = {arXiv preprint arXiv:2212.04088},
  year = {2023}
}

@inproceedings{dalal2024plan,
  title={Plan-seq-learn: Language model guided rl for solving long horizon robotics tasks},
  author={Dalal, Murtaza and Chiruvolu, Tarun and Chaplot, Devendra and Salakhutdinov, Ruslan},
  booktitle={International Conference on Learning Representations},
  volume={2024},
  pages={11087--11115},
  year={2024}
}

@article{zhang2025llm,
  title={LLM-GROP: Visually grounded robot task and motion planning with large language models},
  author={Zhang, Xiaohan and Ding, Yan and Hayamizu, Yohei and Altaweel, Zainab and Zhu, Yifeng and Zhu, Yuke and Stone, Peter and Paxton, Chris and Zhang, Shiqi},
  journal={The International Journal of Robotics Research},
  pages={02783649251378196},
  year={2025},
  publisher={SAGE Publications Sage UK: London, England}
}

@inproceedings{chen2024autotamp,
  title={Autotamp: Autoregressive task and motion planning with llms as translators and checkers},
  author={Chen, Yongchao and Arkin, Jacob and Dawson, Charles and Zhang, Yang and Roy, Nicholas and Fan, Chuchu},
  booktitle={2024 IEEE International conference on robotics and automation (ICRA)},
  pages={6695--6702},
  year={2024},
  organization={IEEE}
}

@inproceedings{wang2024llm3,
  title={Llmˆ 3: Large language model-based task and motion planning with motion failure reasoning},
  author={Wang, Shu and Han, Muzhi and Jiao, Ziyuan and Zhang, Zeyu and Wu, Ying Nian and Zhu, Song-Chun and Liu, Hangxin},
  booktitle={2024 IEEE/RSJ International Conference on Intelligent Robots and Systems (IROS)},
  pages={12086--12092},
  year={2024},
  organization={IEEE}
}

@article{wake2024gpt4v,
  author = {Wake, Naoki and Kanehira, Atsushi and Sasabuchi, Kazuhiro and Takamatsu, Jun and Ikeuchi, Katsushi},
  title = {{GPT-4V(ision)} for Robotics: Multimodal Task Planning from Human Demonstration},
  journal = {IEEE Robotics and Automation Letters},
  volume = {9},
  number = {11},
  pages = {10567--10574},
  year = {2024}
}

@inproceedings{brohan2023can,
  title={Do as i can, not as i say: Grounding language in robotic affordances},
  author={Brohan, Anthony and Chebotar, Yevgen and Finn, Chelsea and Hausman, Karol and Herzog, Alexander and Ho, Daniel and Ibarz, Julian and Irpan, Alex and Jang, Eric and Julian, Ryan and others},
  booktitle={Conference on robot learning},
  pages={287--318},
  year={2023},
  organization={Pmlr}
}

@misc{yao2023react,
  author = {Yao, Shunyu and Zhao, Jeffrey and Yu, Dian and Du, Nan and Shafran, Izhak and Narasimhan, Karthik and Cao, Yuan},
  title = {{ReAct}: Synergizing Reasoning and Acting in Language Models},
  howpublished = {arXiv preprint arXiv:2210.03629},
  year = {2023}
}

@inproceedings{lin2023attention,
  title={Attention for robot touch: Tactile saliency prediction for robust sim-to-real tactile control},
  author={Lin, Yijiong and Comi, Mauro and Church, Alex and Zhang, Dandan and Lepora, Nathan F},
  booktitle={2023 IEEE/RSJ International Conference on Intelligent Robots and Systems (IROS)},
  pages={10806--10812},
  year={2023},
  organization={IEEE}
}

@misc{zhu2023pourme,
  author = {Zhu, Feiya and Hu, Shuo and Leng, Letian and Bartsch, Alison and George, Abraham and Farimani, Amir Barati},
  title = {Pour Me a Drink: Robotic Precision Pouring Carbonated Beverages into Transparent Containers},
  howpublished = {arXiv preprint arXiv:2309.08892},
  year = {2023}
}

@inproceedings{zitkovich2023rt2,
  author = {Zitkovich, Brianna and Yu, Tianhe and Xu, Sichun and Xu, Peng and Xiao, Ted and Xia, Fei and others},
  title = {{RT-2}: Vision-Language-Action Models Transfer Web Knowledge to Robotic Control},
  booktitle = {Conference on Robot Learning (CoRL)},
  pages = {2165--2183},
  year = {2023}
}

@article{ai2025dynamics,
  author = {Ai, Bo and Tian, Stephen and Shi, Haochen and Wang, Yixuan and Pfaff, Tobias and Tan, Cheston and Christensen, Henrik I. and Su, Hao and Wu, Jiajun and Li, Yunzhu},
  title = {A Review of Learning-Based Dynamics Models for Robotic Manipulation},
  journal = {Science Robotics},
  volume = {10},
  number = {106},
  pages = {eadt1497},
  year = {2025},
  doi = {10.1126/scirobotics.adt1497}
}

@inproceedings{goerner2019mtc,
  author = {G{\"o}rner, Michael and Haschke, Robert and Ritter, Helge and Zhang, Jianwei},
  title = {{MoveIt!} {Task} {Constructor} for Task-Level Motion Planning},
  booktitle = {IEEE International Conference on Robotics and Automation (ICRA)},
  pages = {190--196},
  year = {2019},
  doi = {10.1109/ICRA.2019.8793898}
}

@misc{jocher2023yolov8,
  author = {Jocher, Glenn and Chaurasia, Ayush and Qiu, Jing},
  title = {{Ultralytics} {YOLOv8}},
  howpublished = {\url{https://github.com/ultralytics/ultralytics}},
  note = {Version 8.0.0; accessed 15 Aug 2026},
  year = {2023}
}

\end{document}